\documentclass{article}

\usepackage{microtype}
\usepackage{graphicx}
\usepackage{subfigure}
\usepackage{booktabs} 
\usepackage{algpseudocode}

\usepackage{hyperref}

\usepackage[accepted]{icml2023}

\usepackage{amsmath}
\usepackage{amssymb}
\usepackage{mathtools}
\usepackage{amsthm}
\usepackage{mathrsfs}
\usepackage{xcolor}

\usepackage[capitalize,noabbrev]{cleveref}

\theoremstyle{plain}

\theoremstyle{definition}

\theoremstyle{remark}

\usepackage[textsize=tiny]{todonotes}

\icmltitlerunning{Increasing Active Learning Efficiency with Unsupervised Learning in Anomaly Detection}

\begin{document}

\twocolumn[
\icmltitle{Unsupervised Learning of Distributional Properties can Supplement Human Labeling and Increase Active Learning Efficiency in Anomaly Detection}




\icmlsetsymbol{equal}{*}

\begin{icmlauthorlist}
\icmlauthor{Jaturong Kongmanee}{yyy}
\icmlauthor{Mark Chignell}{yyy}
\icmlauthor{Khilan Jerath}{sss}
\icmlauthor{Abhay Raman}{sss}
\end{icmlauthorlist}

\icmlaffiliation{yyy}{University of Toronto}
\icmlaffiliation{sss}{Sun Life Financial}

\icmlcorrespondingauthor{Jaturong Kongmanee}{jaturong.kongmanee@mail.utoronto.ca}

\icmlkeywords{Machine Learning, ICML}

\vskip 0.3in
]



\printAffiliationsAndNotice{}  

\begin{abstract}
Exfiltration of data via email is a serious cybersecurity threat for many organizations. \textcolor{black}{Detecting data exfiltration (anomaly) patterns typically requires labeling, most often done by a human annotator, to reduce the high number of false alarms.} Active Learning (AL) is a promising approach for labeling data efficiently, but it needs to choose an efficient order in which cases are to be labeled, and there are uncertainties as to what scoring procedure should be used to prioritize cases for labeling, \textcolor{black}{especially when detecting rare cases of interest is crucial}.  We propose an adaptive AL sampling strategy that leverages the underlying prior data distribution, as well as model uncertainty, to produce batches of cases to be labeled that contain instances of rare anomalies. We show that (1) the classifier benefits from a batch of representative and informative instances of both normal and anomalous examples, (2) unsupervised anomaly detection plays a useful role in building the classifier in the early stages of training when relatively little labeling has been done thus far. Our approach to AL for anomaly detection outperformed existing AL approaches on three highly unbalanced UCI benchmarks and on one real-world redacted email data set.
\end{abstract}
\vspace*{-3mm}

\begin{figure}[t]
\vskip 0.2in
\begin{center}
\centerline{
\includegraphics[width=.33\textwidth,height=!]{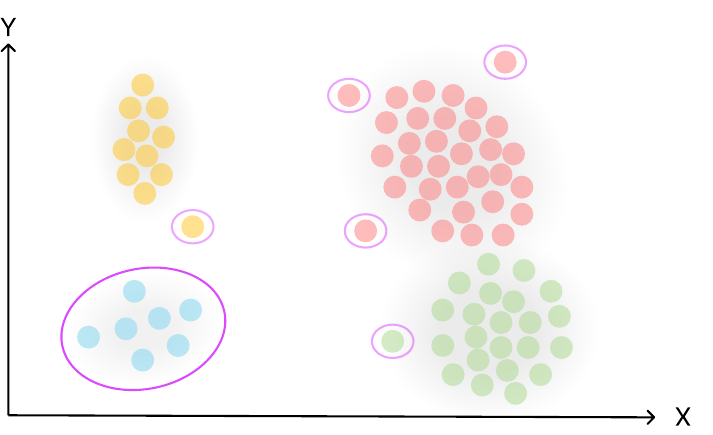}}
\caption{A simple example of anomalies (highlighted with ellipses) in a two-dimensional data set. The large ellipse shows a well-defined group of anomalies, while the ellipses around single points highlight  anomalies lying in low-density regions.}
\label{fig:anomalies-examples}
\end{center}
\vskip -0.2in
\vspace*{-4mm}
\end{figure}

\section{Introduction}
\label{introduction}
Data exfiltration is an unauthorized process of transferring an individual’s or organization’s sensitive data  outside an organization’s perimeter. Exfiltrating data via email is an often-used method and is a serious cybersecurity threat for many organizations, irrespective of whether carried out by organized crime, commercial competitors, external bad actors, or careless or malicious insiders. Sensitive data can be transmitted as plain text in an email body, or attached as a file. There are several solutions available for combating data exfiltration, but they come with their own shortcomings. For example, if  email domains are blacklisted, a determined insider could easily circumvent this by setting up accounts with different domains. Securing email gateways (SEGs) may be effective in blocking phishing emails; however, they can’t stop all spear phishing emails, targeted phishing attacks using social engineering to impersonate trustworthy insiders to trick them into revealing login credentials, installing malware, or stealing data. Rule-Based solutions using ``if-then'' statements and regular expressions to look for data exfiltration signals are impossible to maintain because patterns and sensitivity in data change over time.

In cases where defense at the perimeter is insufficient (as is the case if malicious acts are able to compromise accounts) methods are needed to identify activities such as email data exfiltration (the anomalous instance) that may be carried out using compromised accounts. \textcolor{black}{Anomalous patterns are dynamic in nature and the current notion of normal patterns might not be representative in the future \cite{chandola2009anomaly, hodge2004survey}. Thus, defining a precise decision boundary between normal and anomalous patterns is extremely difficult and is domain-specific. In practice, it is not known if dense regions consist of only normal examples and anomalous examples are those residing in low-density regions near the decision boundary. Moreover, it is also possible that anomalous examples potentially reside in clusters occupying low-density regions (e.g., the circled cluster of points shown in the lower left of \cref{fig:anomalies-examples}).}

Detecting anomalies typically requires labeling, most often done by a human annotator, \textcolor{black}{to build a classifier that can capture evolving anomalous patterns}. However, the labeling process tends to be expensive both in terms of time and cost, and active learning (AL) methods are used to more efficiently take human knowledge into account. AL is a branch of machine learning where the key idea is to sample a small proportion of the data and obtain labels for that sample from a human annotator.

In practice, fully labeled data may not be required since good model performance is often obtained when models are trained on a well-selected subset of the data. Considerable research has demonstrated that AL can produce more efficient labeling of subsets. For instance, \cite{settles2011theories, settles2009active} used AL to maintain  model performance while reducing the size of the labeled training set. Early AL methods have generally assumed that prior class probabilities are similar (balanced classes), which is not realistic in anomaly detection where the proportion of anomalies is extremely low. Sampling strategies based on model uncertainty are widely used but result in an over-reliance on cases near the decision boundary, where there is a danger that the human judge may be no more confident about her labels than the model is about its predictions.

AL is also influenced by the cold start problem \cite{houlsby2014cold,he2009learning,konyushkova2017learning,gao2020consistency} which potentially limits performance for uncertainty-based sampling when the initial training set is limited. One issue of particular concern is the likelihood of sampling biases influencing the model when there are few labels to guide it, where the model may ignore some regions of the sample spaces or even completely overlook certain classes.

Unsupervised ML anomaly detection techniques do not suffer from the cold start problem because they leverage the underlying data distribution rather than labels. Unsupervised ML anomaly detection techniques can improve the detection of new patterns of anomalous and rare examples, but labeling (explicit supervision) is still required in most cases to reduce the high number of false alarms that might otherwise occur.

In this paper, we ask the question, can we combine unsupervised and supervised methods to increase the efficiency of AL and to reduce the amount of human labeling effort required while still achieving a reasonable level of anomaly detection performance? Our answer to this question focuses on enhancing the sampling strategy for AL. We show empirically that the enhanced sampling method outperforms three baseline methods in terms of the area under the precision-recall curve (PRAUC) and anomaly detection rate on each of four data sets (three highly unbalanced UCI data sets and one redacted email provided by a financial company).

\begin{figure}
\centering
{\hfill
    \subfigure[An approach preferring informative instances]{\includegraphics[width=0.2\textwidth]{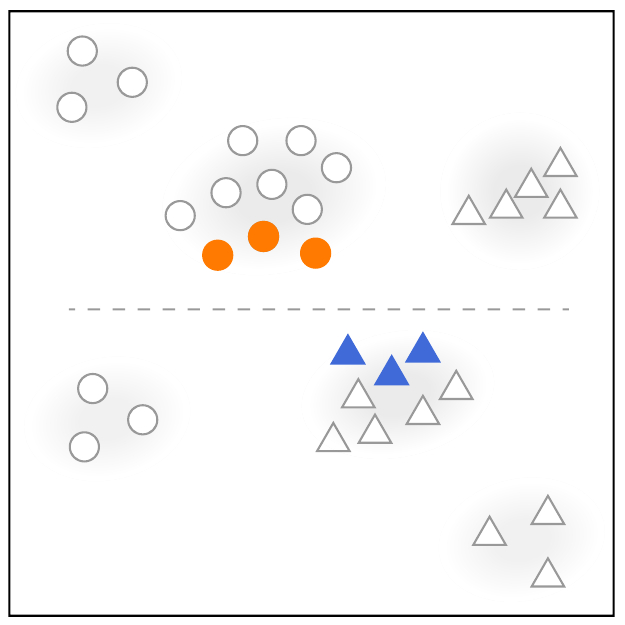}}
    \hfill
    \subfigure[An approach preferring representative instances]{\includegraphics[width=0.2\textwidth]{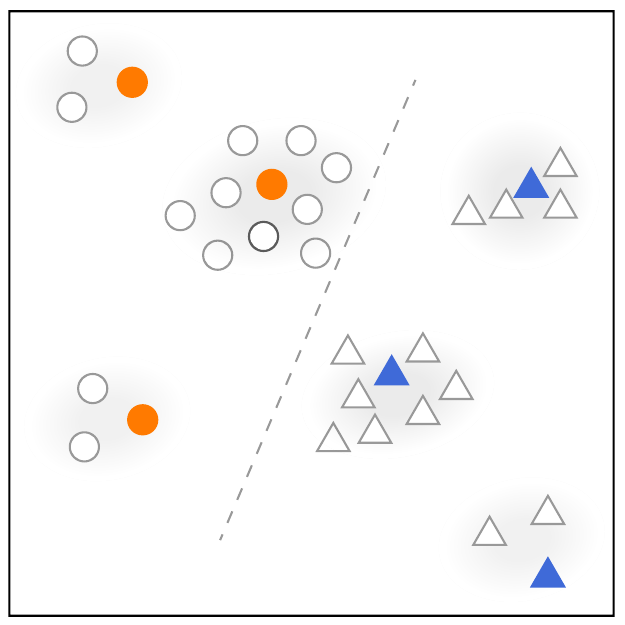}}
\hfill}

{\hfill
    \subfigure[Balancing informativeness and representativeness]{\includegraphics[width=0.2\textwidth]{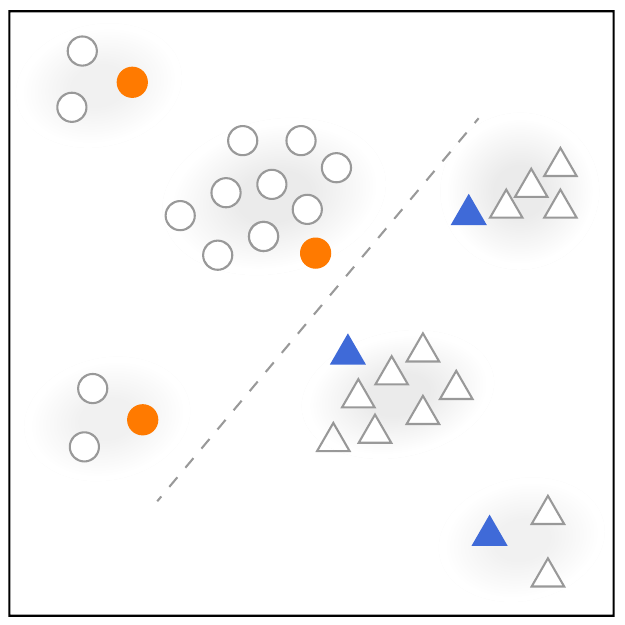}}
    \hfill
    \subfigure[Optimal decision boundary trained on the entire data set]{\includegraphics[width=0.2\textwidth]{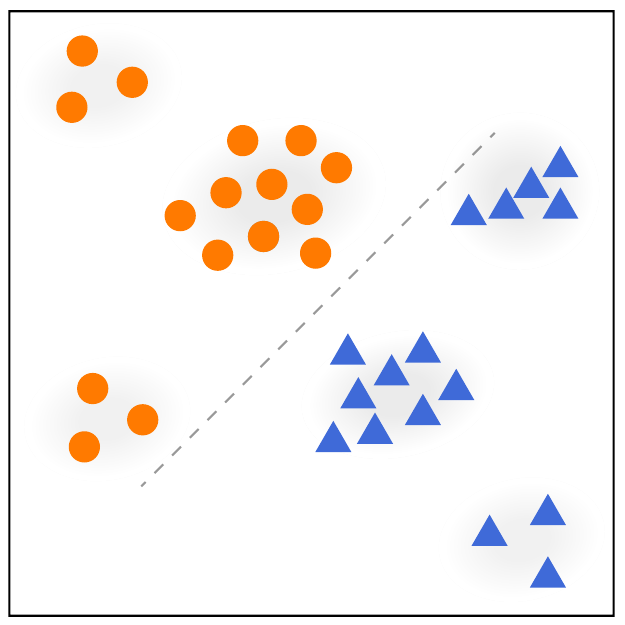}}
\hfill}
\vspace*{-1mm}
\caption{A conceptual illustration of sampling instances of linearly separable data. The white circles and triangles represent unlabeled samples. The orange circles and the blue triangles are labeled as positive and negative, respectively.}
\label{fig:decision-boundary_our-method}
\vspace*{-4mm}
\end{figure}

\section{Related Work}
\label{related_work}


Several efforts have been made to improve sampling strategies in AL. Sampling strategies can be measured based on informativeness, representativeness, or a combination of both. Informativeness refers to the extent to which querying a sample can reduce the model uncertainty. In contrast, representativeness measures how well a sample represents the underlying distribution of unlabeled data \cite{settles2009active}. Sampling the most informative instances has been used extensively, with strategies including query-by-committee \cite{dagan1995committee, freund1997selective, seung1992query}, uncertainty sampling \cite{balcan2007margin, lewis1994heterogeneous, lewis1995sequential, tong2001support}, and optimal experimental design \cite{flaherty2005robust, yu2006active}. The main disadvantage of these strategies is that they ignore the prior data distribution, which can be useful for AL.  The selection of query samples in the initial rounds of AL is based on only a few labeled examples, and can lead to sample bias if the distributional properties of the data are ignored, \textcolor{black}{as shown in Figure \ref{fig:decision-boundary_our-method}a}. This problem is especially noticeable when dealing with data that is highly unbalanced. When a class is rare, its representative cases in that class  may be overlooked because the data distribution of that class cannot be estimated with the relatively few instances sampled. Consequently, potential anomaly examples residing in classes associated with low-density regions may be ignored.

When sampling using an unsupervised approach,  representativeness measures utilize the cluster structure of unlabeled data and focus on selecting the most suitable instances to represent the unlabeled data \cite{nguyen2004active, dasgupta2008hierarchical}. Locally linear reconstructions are used to identify the data samples that adequately reconstruct the entire data set \cite{zhang2011active}. Without utilizing classification uncertainty (since labels are not used), the effectiveness of this approach is highly dependent on the performance of clustering results. \textcolor{black}{As shown in Figure \ref{fig:decision-boundary_our-method}b, the representative sampling selects instances lying at the centers of clusters and can approximate accurate decision boundaries, but many queries are required.} In practice, unsupervised methods need to be supplemented with labeling (supervision) at some point so that the model can  converge to a sufficient level of classification performance. Previous work reported by \cite{huang2010active, ebert2012ralf, kremer2014active} shows that using only one sampling strategy for AL may lead to a reduction in performance.

Early AL algorithms tried to find the optimal query examples by combining informativeness and representativeness measures. \textcolor{black}{In Figure \ref{fig:decision-boundary_our-method}c, balancing both strategies potentially yields a decision boundary close to the optimal (Figure \ref{fig:decision-boundary_our-method}d) with fewer labels.} In \cite{xu2003representative}, the authors proposed a sampling strategy that performs clustering on the instances that are near a decision boundary. One limitation of this approach is its inability to exploit unlabeled examples that are more distant from the decision boundary. \cite{thrun1991active} used an approach that switched randomly between uncertainty sampling and random sampling. \cite{nguyen2004active} dynamically balanced uncertainty and the density of instances using a sampling strategy that pre-clustered data with the k-medoids algorithm. However, the method developed in \cite{nguyen2004active} does not account for unbalanced classes, and the density estimation for each data point is limited to only the current set of clusters. \cite{pelleg2004active, stokes2008aladin} proposed using a fixed combination of low likelihood and high uncertainty criteria for anomaly detection.

Ideally, AL sampling methods should adapt to the amount of “knowledge” that an ML model has about the distribution of cases, and about the relationship between the type of label and the position of instances in the feature space. Non-adaptive sampling criteria (such as those mentioned in the preceding paragraph) do not adjust the scoring criteria as the number of labeling samples increases and learning progresses. For instance, the model may waste effort by sampling near the current decision boundary, where there is often a high degree of uncertainty in the labels and where labeling may add little additional value/information. While fully automated models have achieved some success, they lack flexibility in terms of possible time-varying trade-off between an unsupervised approach (useful when there are only a few labels) and a supervised approach (likely better when the model is better trained). Given that there is this trade-off, it would likely be useful to allow a human annotator to control the behavior of the sampling strategy, at least in some scenarios.


\section{Method}
\label{method}

Sampling based on informativeness measures selects cases residing in low-density regions near the decision boundary of the current model. This approach will be able to find outlier anomalies, but is not capable of finding anomalies that are located in clusters. Clustering the data can be helpful in two ways. First, the representative samples located at the center of clusters are more significant than others and should be prioritized for labeling. Second, samples within the same cluster are likely to have the same label \cite{blum2001learning}. Thus, sampling based on representativeness measures can identify sufficiently accurate decision boundaries, but many calls to query labels from a human annotator are required. Therefore, we design our sampling strategy to be adaptive with a time-varying trade-off. Initially, the strategy is biased towards unsupervised methods (e.g., all ten in a batch of 10 instances for labeling are selected using an unsupervised method). In successive sampling rounds,  there is increasing use of supervised methods.  In the formulation used here (see section \ref{subsub:adaptive-sampling}), parameters determine how quickly the transition from predominantly unsupervised to predominantly supervised sampling occurs during successive rounds. By parameterizing the scoring criterion, the human annotator is given control of the parameter settings that specify the trade-off between the number of unsupervised and supervised instances in each AL round. In the remainder of this section, we first introduce the batch mode AL approach and then list the heuristics that guide a potentially more effective AL approach.

\subsection{Active Learning}
\label{sub:active-learning}
In this section, we formally describe the AL approach that will be used. Given a classifier $f(\textbf{x}; \theta)$, unlabeled samples $\mathcal{U}$, a labeled training set $\mathcal{L}$, and input $\textbf{x} \in \mathcal{U}$, a sampling strategy $\phi(\textbf{x}, f(\textbf{x}; \theta))$ is a function of $\textbf{x}$ and $f(\cdot)$ that the AL uses to select samples for labeling:

{\small
\vspace*{-5mm}
\begin{align}\label{eq: acuisition-func}
    &\textbf{x}^{*} = \arg \max_{\textbf{x} \in \textit{U}} \phi(\textbf{x}, f(\textbf{x}; \theta)),
\end{align}
\vspace*{-6mm}
}

A batch mode AL selects batches of $b$ instances at a time for human annotation to obtain an accurate model at a lower labeling cost than regular supervised learning. The standard AL procedure is as follows:
\vspace*{-2mm}
\begin{enumerate}
    \item Select a set of unlabeled instances $\mathcal{L} \subset \mathcal{U}$ for labeling.
    \item Train a classifier $f(\textbf{x}; \theta)$ with $\mathcal{L}$
    \item Select $\textbf{x}^* \in \mathcal{U} \setminus \mathcal{L}$ for labels using $\phi(\cdot)$
    \item Assign labels $\textbf{y}^*$ to $\textbf{x}^*$ and update pools of labeled and unlabeled samples
    \item Repeat steps $2 - 5$ until the classifier's performance is achieved or a number of iterations have reached a predefined number.
\end{enumerate}

\subsubsection{Representative Sampling}
\label{subsub:representative-sampling}
Our approach to AL includes representative sampling, where the goal is to learn the underlying prior data distribution of the unlabeled data and to select batches of representative samples in the early stage of AL. Mixture density estimation is used to delineate important regions (including anomalous examples residing in clusters with low-density regions) of the sample space to avoid sampling biases that may occur where a small number of labeled instances are not representative of the overall data set. Thus, we propose to use unsupervised analysis of the multivariate data distribution to reduce bias in the early stages of training. This is done by representing density variations in the space that guide anomaly detection when labeled instances are rare.

In this work, we represent the underlying data distribution presumably generated by a mixture model in terms of the Gaussian Mixture Model (GMM). The GMM is a parametric probability density function represented as a weighted sum of Gaussian component densities \cite{dempster1977maximum, mclachlan1988mixture}. Compared to other clustering methods, such as K-Means, a GMM provides statistical inferences concerning the underlying distributions that can be used later to determine the degree of anomaly of samples \cite{aggarwal2017introduction, wang2019progress, yang2021generalized}. We expect different Gaussian components in the mixture to learn different distributions that correspond to a variety of data patterns.
The parameters of the mixture model are estimated by maximum likelihood estimate (MLE) via the Expectation-Maximization algorithm \cite{dempster1977maximum}. Our method works as follows:

\begin{enumerate}
    \item Identify the number of mixture components ($\mathcal{K}$) corresponding to which of the alternative formulations has the lowest Bayesian Information Criterion (BIC) score \cite{schwarz1978estimating}.
    \item Fit GMM with $\mathcal{K}$ components using EM.
    \item Return $n_{repr}$ centroids with the lowest probability density if $\mathcal{K} \geq n_{repr}$, where $n_{repr}$ is the number of instances being selected in a batch, or else return $n_{repr} + (n_{repr} - \mathcal{K})$, where $(n_{repr} - \mathcal{K})$ is the instances with the lowest likelihood presumably generated by the learned distribution.
\end{enumerate}

Our representative sampling returns centroids, including potential anomaly centroids residing in clusters with low-density regions that can approximate the underlying data distribution. Ranking each sample in order of increasing model likelihood and selecting the most anomalous instances to minimize model variances, can improve refinement of decision boundaries.

\subsubsection{Informative Sampling}
\label{subsub:informative-sampling}
Our sampling approach uses information entropy \cite{shannon1948mathematical} as a measure of uncertainty/informativeness.  This method selects the most informative samples, i.e., the samples that are close to the decision boundary, presumably passing through low-density regions of the marginal data distribution. The least informative samples are those where one of the classes has a high probability (examples far away from the decision boundary). Formally, for $k$-class classification, the information entropy $\mathbb{H}(x)$ of sample $x$ can be defined as $\mathbb{H}(x) = -\sum_{i=1}^{k} P(y_i|x)\cdot log P(y_i|x)$, where $P(y_i|x)$ is the probability that the current sample $x$ is predicted to be class $y_i$. The greater the entropy of the sample, the greater its uncertainty, which we refer to as \textit{Max Entropy}. However, a batch mode AL strategy that selects multiple informative samples each time might result in samples that are very similar, providing little information. Thus, the selected batch should be informative for the model, while being diverse enough to minimize redundancy between sampled instances. Our method operates in the following way:

\begin{enumerate}
    \item Select the top $\frac{n_{info} \times 100}{b}\%$ most informative instances from the pool of unlabeled data $\mathcal{U}$, where $n_{info}$ is the number of instances being selected in a batch.
    \item Apply K-Means clustering to all informative instances obtained from the previous step to identify $n_{info}$ groups. The k-means++ seeding algorithm \cite{arthur2007k} is used to promote diversity among these informative instances.
    \item Return $n_{info}$ instances that are closest to the cluster centroids for human labeling.
\end{enumerate}

In summary, our proposed informative sampling heuristic avoids the selection of redundant instances and concentrates on the most important informative instances in selecting samples.

\subsubsection{Adaptive Sampling}
\label{subsub:adaptive-sampling}

Our approach aims to combine representative and informative samplings as a function of AL iterations. The proposed approach prioritizes representative sampling in the initial phase and linearly\footnote{We also experimented with exponential and polynomial, but we found linearly transitioning between two criteria worked best consistently across all experiments.} balances both criteria until informative sampling becomes dominant. This ensures there is always a mixture of the two criteria in the early AL stage. Since  human experience is a valuable resource and should be incorporated into solving a problem, we allow a human annotator to control the behavior of sampling strategies to improve the model performance. The \textit{balancing function} has the form:

{\small
\vspace*{-4mm}
\begin{align}
    \label{eq: balance_rate}
    &\alpha(t, b, c, T_1, T_2) = \begin{dcases}
        b, 0 & t< T_1\\
        \underbrace{b - \mathcal{B(\cdot)}}_\text{$n_{repr}$}, \underbrace{\mathcal{B(\cdot)}}_\text{$n_{info}$} & T_1 \leq t < T_2; t=t-T_1 \\
        0, b & T_2\leq t\\
    \end{dcases}
\end{align}
\vspace*{-5mm}
}

where \textit{t} is the AL iteration, \textit{b} is the batch size, $\textit{c} \in [0, 1]$ is a human annotator's confidence level for her initial classifier, $T_1$ is the iteration to start balancing (i.e., adding in informative samples of some cases), $T_2$ is the stopping iteration (i.e., using only the informative sampling), $\mathcal{B(\cdot)}$ is $\mathbf{mod}(t+\lceil{b*c}\rceil, b)$, and the function returns two values: $n_{repr}$ and $n_{info}$. $n_{repr}$ is the number of instances selected through the representative sampling, while $n_{info}$ is based on informative sampling, where the sum of these values equals $b$. Pseudo-code for the proposed algorithm is given in Alg. \ref{alg: adaptive-AL}.

\begin{algorithm}[t]
\caption{Adaptive AL Sampling Strategy}
\label{alg: adaptive-AL}
\begin{algorithmic}[1]
\Require unlabeled data set $\mathcal{U}$, labeled set $\mathcal{L}$, batch size $\textit{b}$, initial number of labeled examples $\mathcal{M}$, number of iterations $\mathcal{T}$, classifier $f(\textbf{x};\theta)$, sampling strategy $\phi(\cdot)$, balancing function $\alpha(\cdot)$, iteration to start balancing $T_1$, iteration to stop balancing $T_2$, confidence level $\textit{c}$.

\State Labeled data set $\mathcal{L} \leftarrow \mathcal{M}$ examples drawn uniformly at random from $\mathcal{U}$ along with queried labels.

\State Train an initial classifier $\theta_0$ on $\mathcal{L}$

\For {$t =  1,2,\ldots, \mathcal{T}$}
    \State $n_{repr}, n_{info} \gets$ $\alpha(\textit{t}, \textit{b}, \textit{c}, T_1, T_2)$ \algorithmiccomment{See \ref{eq: balance_rate}}
    \State $\mathcal{X}_{repr} \gets \textsf{Repr}$$(n_{repr}, \phi(\theta_{t-1}, \mathcal{U}))$  \algorithmiccomment{Section \ref{subsub:representative-sampling}}
    \State $\mathcal{X}_{info} \gets \textsf{Info}$$(n_{info}, \phi(\theta_{t-1}, \mathcal{U}))$  \algorithmiccomment{Section \ref{subsub:informative-sampling}}
    \State $\hat{\mathcal{X}} \gets \{\mathcal{X}_{repr}, \mathcal{X}_{info} \}$
    \State Query labels for $\hat{\mathcal{X}}$
    \State $\mathcal{L} \gets \mathcal{L} \cup \hat{\mathcal{X}}$
    \State $\mathcal{U} \gets \mathcal{U} \setminus  \hat{\mathcal{X}}$
    \State $\theta_{t} \gets$ Train$(\theta_{t-1}, \mathcal{L})$ \algorithmiccomment{Train a classifier on a newly updated training set.}
 \EndFor
\State \textbf{return} The model $\theta_{\mathcal{T}}$

\end{algorithmic}
\end{algorithm}

When there is access to a sufficiently large training set, a human annotator can modify parameter $c$ accordingly.  For instance, setting $c$ to $0.5$ results in a batch that consists of samples selected by both criteria (i.e., a $50/50$ supervised and unsupervised rounds) starting from the first iteration, as opposed to having the unsupervised approach dominating in the first iteration. We hypothesize that switching between supervised and unsupervised training based on the amount of labeled instances already available, and model uncertainty associated with the current level of training, will be useful in creating more efficient AL.

\begin{figure*}
\centering
\subfigure[Abalone]{\includegraphics[width=0.5\columnwidth]{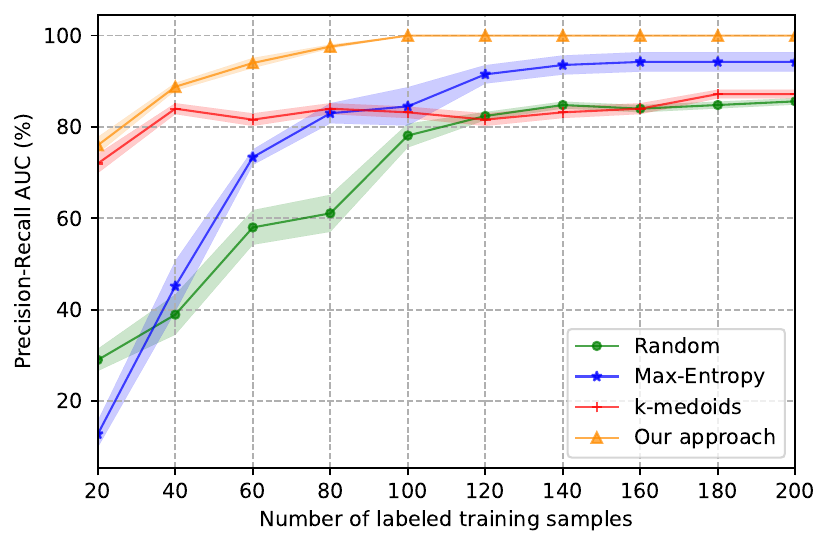}}
\hfill
\subfigure[ANN-Thyroid]{\includegraphics[width=0.5\columnwidth]{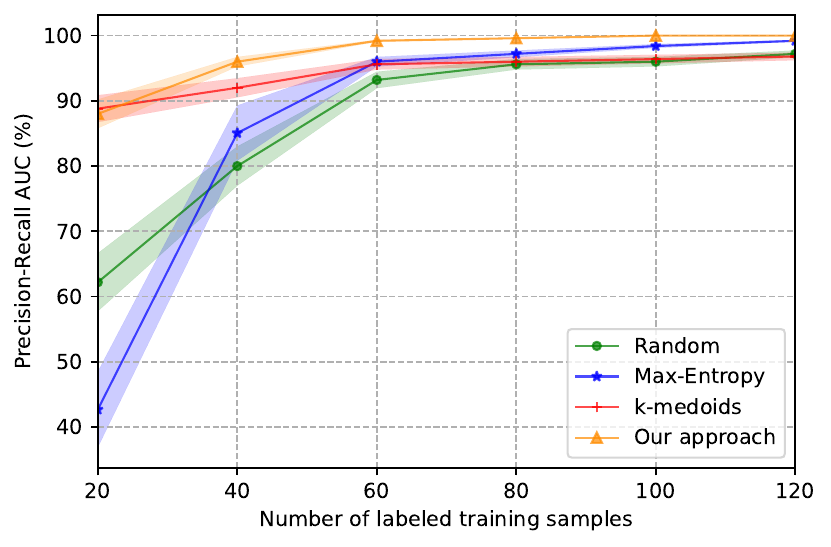}}
\hfill
\subfigure[Cardiotocography]{\includegraphics[width=0.5\columnwidth]{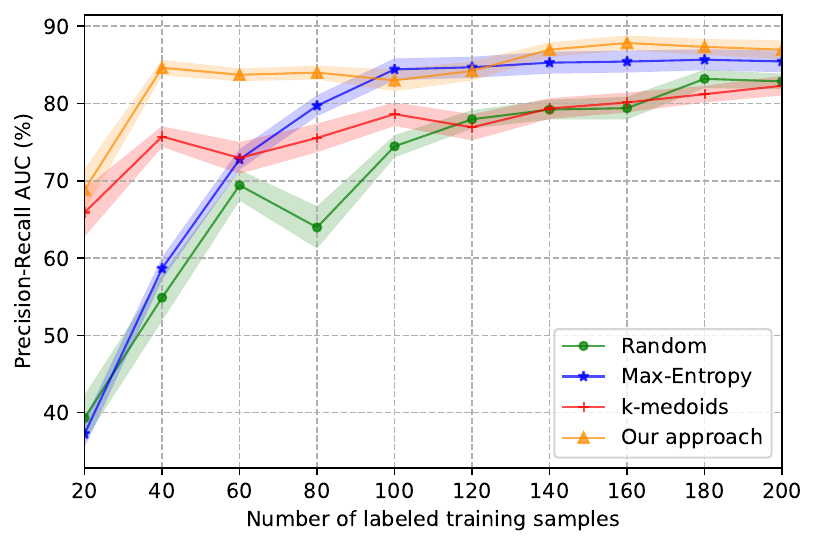}}
\hfill
\subfigure[Redacted Email]{\includegraphics[width=0.5\columnwidth]{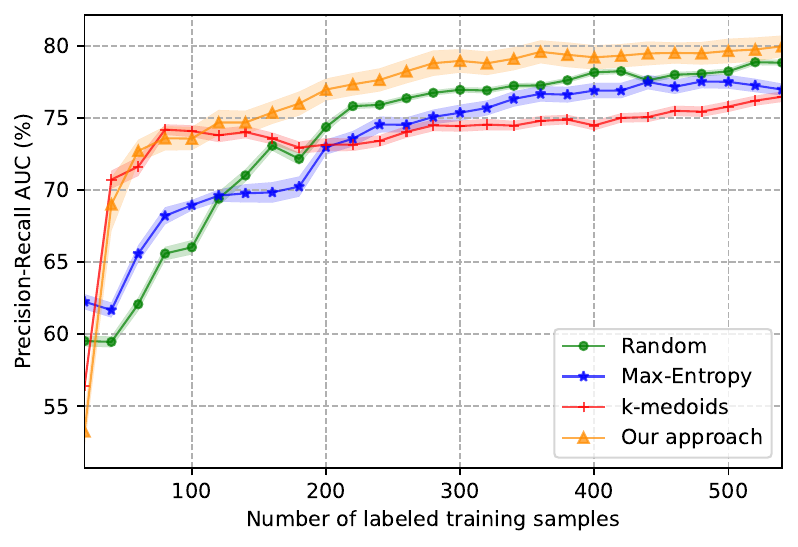}}
\caption{PRAUC, on four different data sets, compared against size of training set (accumulating number of instances sampled).}
\label{fig:PRAUC-performance}
\end{figure*}

\section{Experiments}
\label{experiments}

\begin{table}[ht]
\vspace*{-4mm}
    \caption{Dataset Statistics}
    \label{tab: dataset_table}
    \begin{center}
    \begin{small}
    \begin{sc}
    \resizebox{\columnwidth}{!}{
        \begin{tabular}{lccr}
        \toprule
        Data set & Dimensions & samples & anomalies (\%) \\
        \midrule
        Abalone    & 9 & 1920 & 29(1.50\%) \\
        ANN-Thyroid-1v3 & 21 & 3251 & 73(2.25\%)\\
        Cardiotocography & 22 & 1700 & 45(2.65\%) \\
        Redacted Email & 42 & 672 & 418(62.20\%)\\
        \bottomrule
        \end{tabular}
    }

    \end{sc}
    \end{small}
    \end{center}
    \vskip -0.1in
\end{table}

\subsection{Experimental setup}
\label{sub:experiment-setup}

\textbf{Data sets.} Following \cite{das2016incorporating, zong2018deep}, we evaluate our method on the following highly unbalanced UCI benchmarks \cite{asuncion2007uci} used for anomaly detection: \textit{Abalone}, \textit{Thyroid (ANN-Thyroid)}, \textit{Cardiotocography}, and on one real-world \textit{redacted email} data set (tabular data) provided by a financial service company. The \textit{redacted email} data set had 42 features, including variables such as binary variables that indicate whether certain sensitive terms are present in the subject line or attachment names (full details provided in \cite{wang2023implementing}). A number of anomalies and normal examples in each dataset are shown in Table \ref{tab: dataset_table}. We note that while our method focuses on the highly unbalanced data sets, improving the sampling strategy, in general, will further improve AL for the case of the more balanced data set (i.e., redacted email), as observed in section \ref{sub:experiment-results} that our method benefits from carefully selecting important instances.

\textbf{Baselines and method.} We compared our method with the following baselines: \textbf{i) Random:} The naive baseline of selecting a batch of size $b$ uniformly at random from the unlabeled pool at each round for labeling. This baseline allows us to compare the benefit of AL over passive learning. \textbf{ii) Max Entropy:} A widely used informative sampling strategy baseline that selects a batch of $b$ informative instances according to the entropy of the example’s predictive class probability distribution. For binary classification, max entropy is equivalent to margin sampling and least confident sampling approaches \cite{settles2009active}. \textbf{iii) k-medoids:} A robust-to-noise unsupervised anomaly detection technique that selects $b$ medoids. In contrast to K-Means, k-medoids use data points in a data set as estimates of central location instead of centroids (means), which may not belong to the clusters. Also, k-medoids is less influenced by outliers and noise, making it more robust than K-Means. Previous research by \cite{syarif2012data, agrawal2015survey} shows that k-medoids produces better results than K-Means in detecting novel network anomalies in cybersecurity.

\textbf{Evaluation metrics.} We used a threshold-invariant metric, the area under the precision-recall curve (PRAUC), which is suitable for rare binary events and unaffected by model specificity \cite{davis2006relationship}, and has been shown to be more informative than AUROC score when the classes are highly unbalanced \cite{saito2015precision}. We also plotted a total number of true anomalies discovered as a function of number of queries presented to the human annotator. Ideally the number of true anomalies identified should increase quickly and is thus a measure of the quality of AL performance. Another reason for expecting number of true anomalies to increase quickly is that we want to make efficient use of the human annotator.

\textbf{Implementation details.} Unless otherwise specified, in all experiments, we use Support Vector Machines (SVMs) with a Radial Basis Function (RBF) kernel as a classifier due to its well-understood theoretically \cite{kremer2014active}. We calibrate the probabilities of a classifier using Platt scaling \cite{platt1999probabilistic}. We divided each data set into two sets using Stratified Shuffle Split \footnote{\url{https://scikit-learn.org/stable/modules/generated/sklearn.model_selection.StratifiedShuffleSplit.html}} to preserve the same percentage for each class as in the original data set. All sampling strategies were performed on the unlabeled set (80\%), and the effectiveness of the sampling strategies was evaluated after each batch based on the other unseen fixed set (20\%) referred to as the test set. We considered a hard case of AL, where we started with two randomly selected labeled examples per class and set a confidence level $(c=0)$. We set $T_1=0$, $T_2=5$, and evaluated them in a batch mode AL setup with a batch size of $b=20$.  The batch sample size of $20$ had been found to be the maximum that could be implemented with human annotators in a previous survey study by \cite{wang2023implementing} at a large financial services company. All sampling strategies started with the same initial labeled set, unlabeled set, and test set. The experiments were repeated for $50$ independent runs, and mean performance, with $95\%$ confidence intervals, are reported.

\subsection{Experimental results and discussion}
\label{sub:experiment-results}

\textcolor{black}{\textbf{PRAUC performance on the cold start problem.}} In Figure \ref{fig:PRAUC-performance}, we demonstrate the empirical results with four initial labeled instances (two for each class) across all data sets and baselines. Our method outperforms Max Entropy sampling by focusing on learning different distributions that correspond to a variety of data patterns, without overlooking a potential rare class, to more effectively estimate decision boundaries within the early AL stage. It can be seen that higher performance was achieved across data sets using our method, presumably due to a reduction in sampling bias. Compared to k-medoids, our method starts with a competitive level of performance and converges more quickly to a high level of performance.

We hypothesize that our method benefits from the greater use of the proposed informative sampling in later AL rounds.  Max Entropy outperforms k-medoids after a sufficient number of labeled samples are collected. However, unsupervised methods may not converge to sufficiently high levels of performance and even if they do, the labeling costs may be too high. So, our method provides a way to adjust the trade-off between unsupervised and supervised learning so that sampling bias can be reduced in earlier AL rounds (focusing on an unsupervised approach) while greater focus on labeled instances can efficiently enhance model performance in later rounds of AL. Our main contribution in this paper is that we provide a novel way to control the trade-off in AL between exploration of the feature space to avoid sampling bias (unsupervised learning) and learning from labeled instances (supervised learning).

\begin{figure}
\centering
    \subfigure[$\mathcal{M}=4$]{\includegraphics[width=0.235\textwidth]{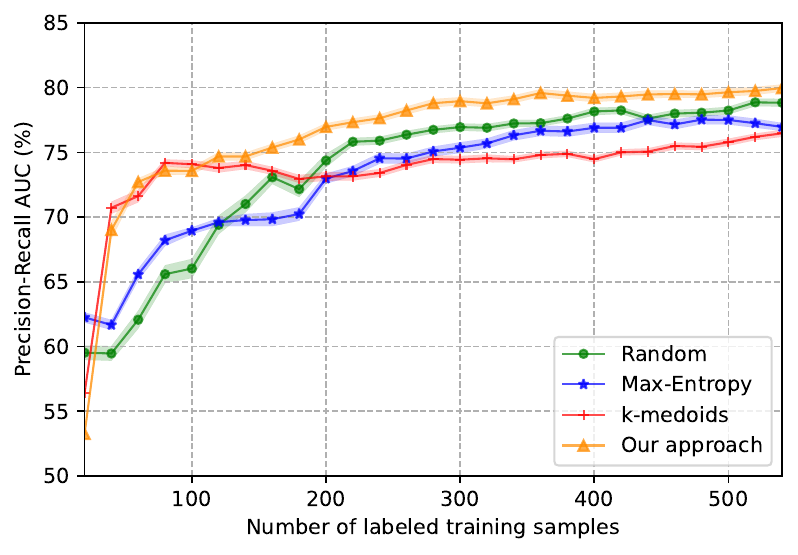}}
    \hfill
    \subfigure[$\mathcal{M}=20$]{\includegraphics[width=0.235\textwidth]{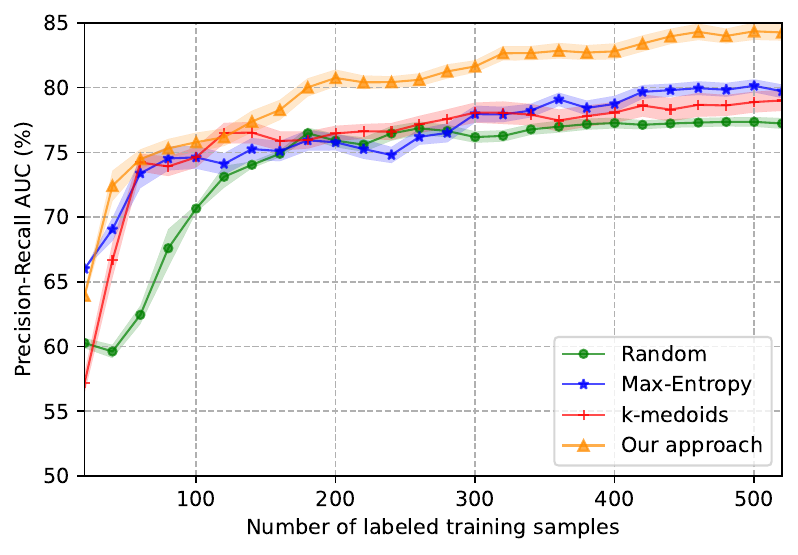}}
\vspace*{-1mm}
\caption{PRAUC on a redacted email test set, compared against the size of the training set for two settings: the amount of the initial labeled set ($\mathcal{M}=4$)  and ($\mathcal{M}=20$).}
\label{fig:cold-start_problem}
\vspace*{-3mm}
\end{figure}

\textcolor{black}{\textbf{Manually varying the trade-off between both sampling strategies.}} We further verify the flexibility and effectiveness of our method on a redacted email dataset. Figure \ref{fig:cold-start_problem}a (identical to Figure 3d but repeated here for comparison purposes) shows that our method can mitigate the effects of sampling bias, as evidenced in Figure \ref{fig:PRAUC-performance}, by initially setting $c$ to $0$ (i.e., starting with the unsupervised learning that explores the data distribution).  As expected, the PRAUC performance in the first iteration is lower than Max Entropy. However, the unsupervised technique used does not suffer from the cold start problem, and outperforms Max Entropy in later iterations. By the seventh iteration (where a total of $140$ instances have been labeled), our method provides a batch that consists of samples purely selected by the proposed informative sampling, which leads to higher performance than k-medoids. Figure \ref{fig:cold-start_problem}b shows the benefit of having access to a sufficiently large training set. In this setting, we adjust parameter $c$ to $0.5$ to obtain a batch of samples selected by both criteria in equal amounts, instead of having the unsupervised approach dominate from the beginning (i.e., $c=0$). Our method closely matches the performance of Max Entropy in early rounds as the initial model has better knowledge about the feature space, demonstrating the benefit of incorporating human knowledge into controlling the behavior of sampling
strategies. The proposed method at the seventh iteration selects non-redundant samples solely based on the informative measures as the amount of labeled examples increases, achieving higher performance than all baselines.

\begin{figure}
\centering
{\hfill
    \subfigure[Abalone]{\includegraphics[width=0.235\textwidth]{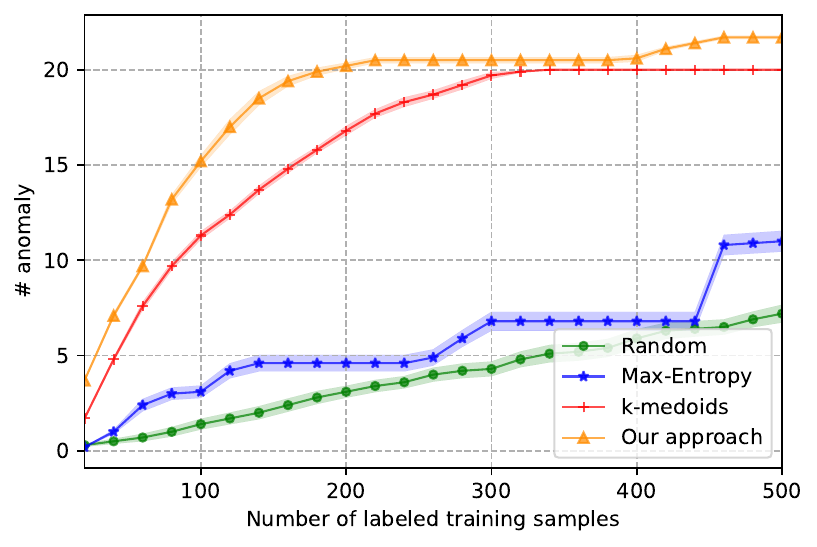}}
    \hfill
    \subfigure[ANN-Thyroid]{\includegraphics[width=0.235\textwidth]{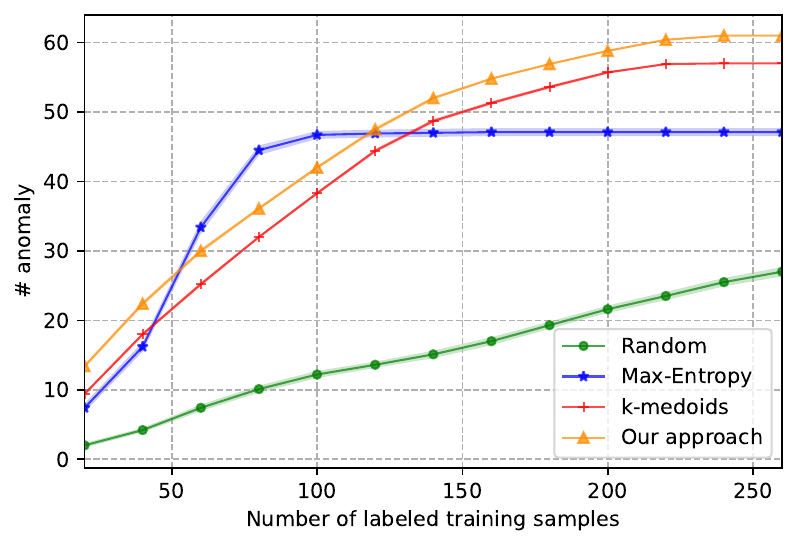}}
\hfill}

{\hfill
    \subfigure[Cardiotocography]{\includegraphics[width=0.235\textwidth]{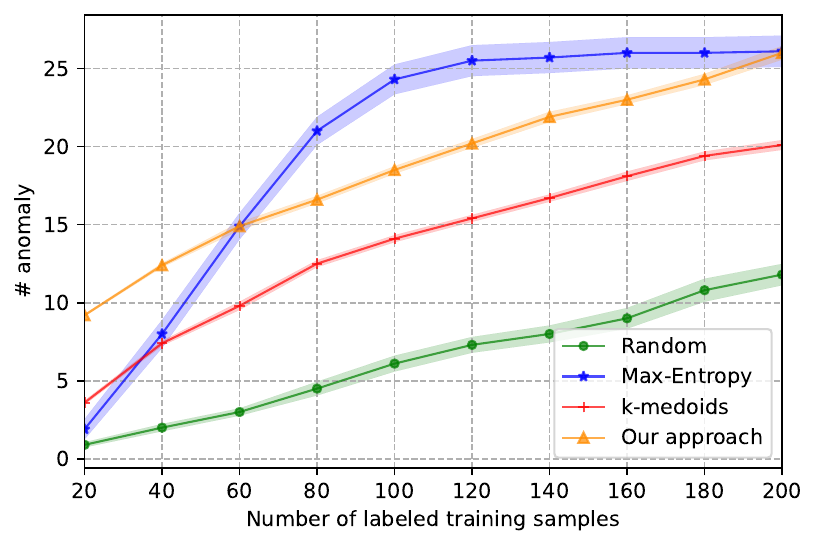}}
    \hfill
    \subfigure[Redacted Email]{\includegraphics[width=0.235\textwidth]{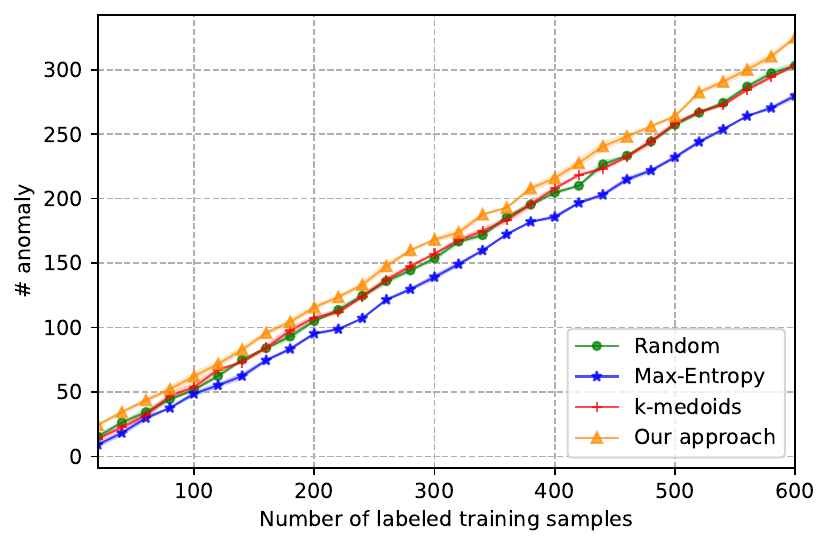}}
\hfill}
\vspace*{-1mm}
\caption{Number of true anomalies discovered on four different data sets, compared against size of training set (accumulating number of instances sampled).}
\label{fig:anomaly-detection}
\vspace*{-3mm}
\end{figure}

\textbf{Anomaly detection rate.} In this experiment, we compare how quickly algorithms can identify anomalous classes in a data set. This will help optimize the use of human annotators' time. The results are illustrated in Figure \ref{fig:anomaly-detection} for our method and the three existing approaches. Our method quickly identifies anomalous samples and is able to include true anomaly examples for human labeling from the first iteration, as opposed to Max Entropy and uniformly sampling approaches. All methods perform equally well for a redacted email data set. We hypothesize this is because the classes in this data set were balanced. Our method exhibits sample-efficient properties by demonstrating performance improvements (Figure \ref{fig:PRAUC-performance}b and \ref{fig:PRAUC-performance}c) while detecting fewer anomalies than Max Entropy in just a few iterations (Figure \ref{fig:anomaly-detection}b and \ref{fig:anomaly-detection}c). We hypothesize that our method prevents the selection of redundant instances and instead focuses on the most important informative instances.

\section{Conclusion}
We have demonstrated that some of the widely used sampling strategies for AL perform poorly in practical scenarios where classes are unbalanced. Our proposed method works well in the presence of highly unbalanced classes and anomalies, as well as when anomalies are frequent. Our simulations show that the method proposed here leads to AL rounds where batches of samples contain instances of rare anomalies. Batches of instances that contain only one class (typically no anomalies when anomalies are rare) will not lead to much new information when cases are labeled. Thus in order to efficiently learn distinctions between anomalies and non-anomalies, there should be examples of anomalies in every batch, more effectively utilizing  human annotator time in the labeling process. Our approach is aimed at increasing the sampling of rare classes, and it is flexible, since we do not assume a particular data distribution, making it applicable to a wide range of data sets. Our approach provides several indicators to assist a human annotator in identifying anomalous data, as well as controlling the behavior of sampling strategy in different settings.

\section*{Acknowledgements} We would like to thank Thanyathorn Thanapatheerakul and Worrawat Engchuan for helpful discussions.

\bibliography{main}
\bibliographystyle{icml2023}

\end{document}